%% file: main.tex
\documentclass[10pt,twocolumn,letterpaper]{article}

\usepackage{iccv}              
\usepackage[accsupp]{axessibility}


\usepackage{amsmath}
\usepackage{multicol}

\usepackage{textcomp}

\usepackage{amsthm}
\usepackage{amssymb}
\usepackage{amsmath}

\usepackage{multirow}
\usepackage{booktabs} 

\usepackage{arydshln}

\usepackage{algorithm}
\usepackage{algorithmic}

\definecolor{iccvblue}{rgb}{0.21,0.49,0.74}
\usepackage[pagebackref,breaklinks,colorlinks,allcolors=iccvblue]{hyperref}

\def\eg{\emph{e.g., }}

\def\wrt{\emph{w.r.t. }}

\def\name{EFTViT}
\newcommand{\bfsection}[1]{\vspace*{0.0mm}\noindent\textbf{#1.}}
\usepackage{xcolor}

\def\paperID{1587} 
\def\confName{ICCV}
\def\confYear{2025}

\title{\name{}: Efficient Federated Training of Vision Transformers with Masked Images on Resource-Constrained Clients}

\author{
Meihan Wu\textsuperscript{1} \quad
Tao Chang\textsuperscript{1} \quad
Cui Miao\textsuperscript{1} \quad
Jie Zhou\textsuperscript{1} \quad
Chun Li\textsuperscript{2} \\
Xiangyu Xu\textsuperscript{3} \quad
Ming Li\textsuperscript{4}\thanks{Corresponding author} \quad
Xiaodong Wang \textsuperscript{1}
\\
\textsuperscript{1}	National University of Defense Technology \quad
\textsuperscript{2} Shenzhen MSU–BIT University
\\
\textsuperscript{3} Xi'an Jiaotong University \\
\textsuperscript{4} Guangdong Laboratory of Artificial Intelligence and Digital Economy (SZ)
}

\begin{document}
\maketitle
\input{0_abstract}    
\input{1_intro}
\input{2_re}
\input{method}
\input{evaluation}
\input{conclusion}

\input{Acknowledgements}

{
    \small
    \bibliographystyle{ieeenat_fullname}
    \bibliography{main}
}

\end{document}

%% file: 0_abstract.tex
\begin{abstract}
Federated learning research has recently shifted from Convolutional Neural Networks (CNNs) to Vision Transformers (ViTs) due to their superior capacity. ViTs training demands higher computational resources due to the lack of 2D inductive biases inherent in CNNs. However, efficient federated training of ViTs on resource-constrained clients remains unexplored in the community. 
In this paper, we propose \name{}, a hierarchical federated framework that leverages masked images to enable efficient, full-parameter training on resource-constrained clients, offering substantial benefits for learning on heterogeneous data. 
In general, we patchify images and randomly mask a portion of the patches, observing that excluding them from training has minimal impact on performance while substantially reducing computation costs and enhancing data content privacy protection. 
Specifically, \name{} comprises a series of lightweight local modules and a larger global module, updated independently on clients and the central server, respectively. The local modules are trained on unmasked image patches, while the global module is trained on intermediate patch features uploaded from the local client, balanced through a proposed median sampling strategy to erase client data distribution privacy. 
We analyze the computational complexity and privacy protection of \name{}. Extensive experiments on popular benchmarks show that \name{} reduces local training computational cost by up to $5.6\times$, cuts local training time by up to $3.1\times$, and achieves up to 2.46\% accuracy improvement compared to existing methods.
\end{abstract}

%% file: 1_intro.tex
\section{Introduction}
\label{sec:intro}


Federated learning (FL)~\cite{mcmahan2017communication} has emerged as a key privacy-preserving machine learning paradigm, with early research oriented computer vision primarily centered on convolutional neural networks (CNNs)~\cite{he2016deep}.
Recently, the focus has increasingly shifted toward Vision Transformers (ViTs) \cite{dosovitskiy2020image}, whose self-attention mechanisms excel at capturing long-range correspondences within images, achieving state-of-the-art performance across visual problems, \eg object recognition~\cite{dosovitskiy2020image}, detection~\cite{he2022masked,dai2021up}, and semantic segmentation~\cite{zheng2021rethinking}.
However, federated training ViTs generally incurs significantly higher computational costs and longer training times due to the lack of spatial inductive biases within images \cite{touvron2021training, chen2022dearkd}, making it prohibitively challenging for resource-constrained FL.



\begin{figure}[!t]
\centering 
    \subfloat[Previous CNN Method ]{\label{fig:p2}\includegraphics[width = 0.48\textwidth]{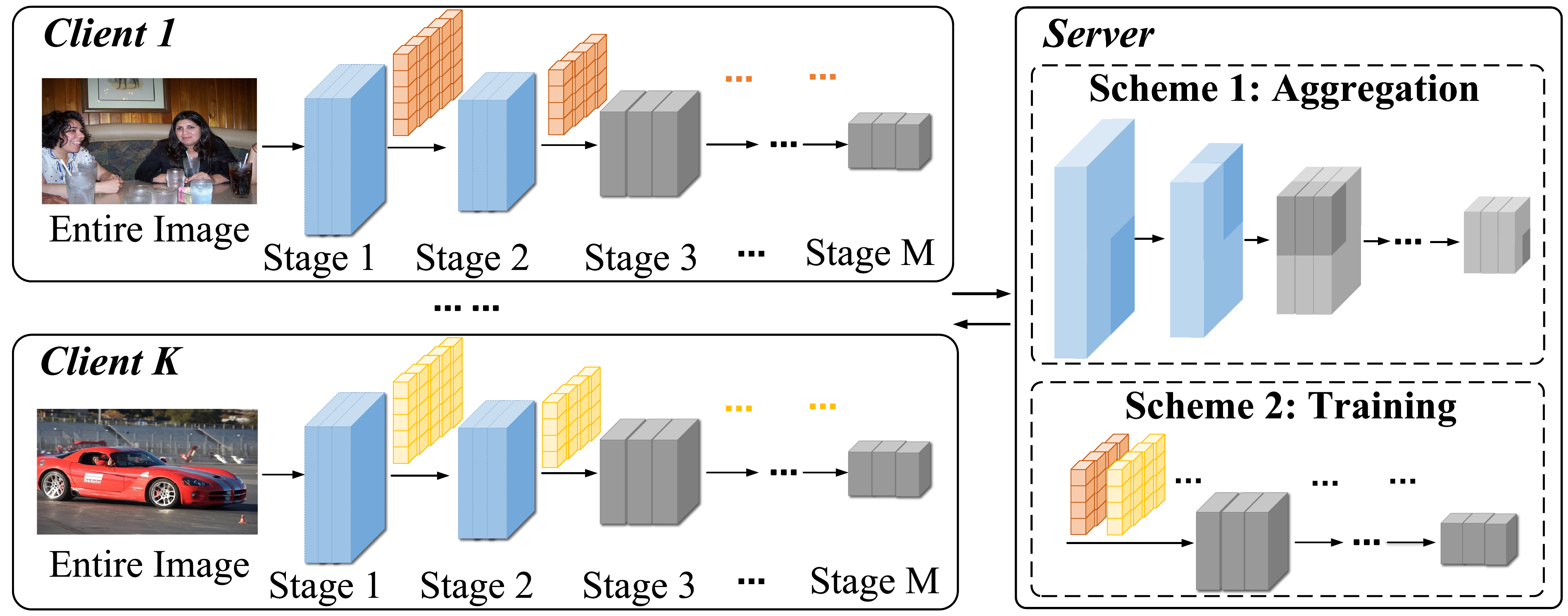}}
    \hfill
    \subfloat[Our ViT Method]{\label{fig:p1}\includegraphics[width = 0.48\textwidth]{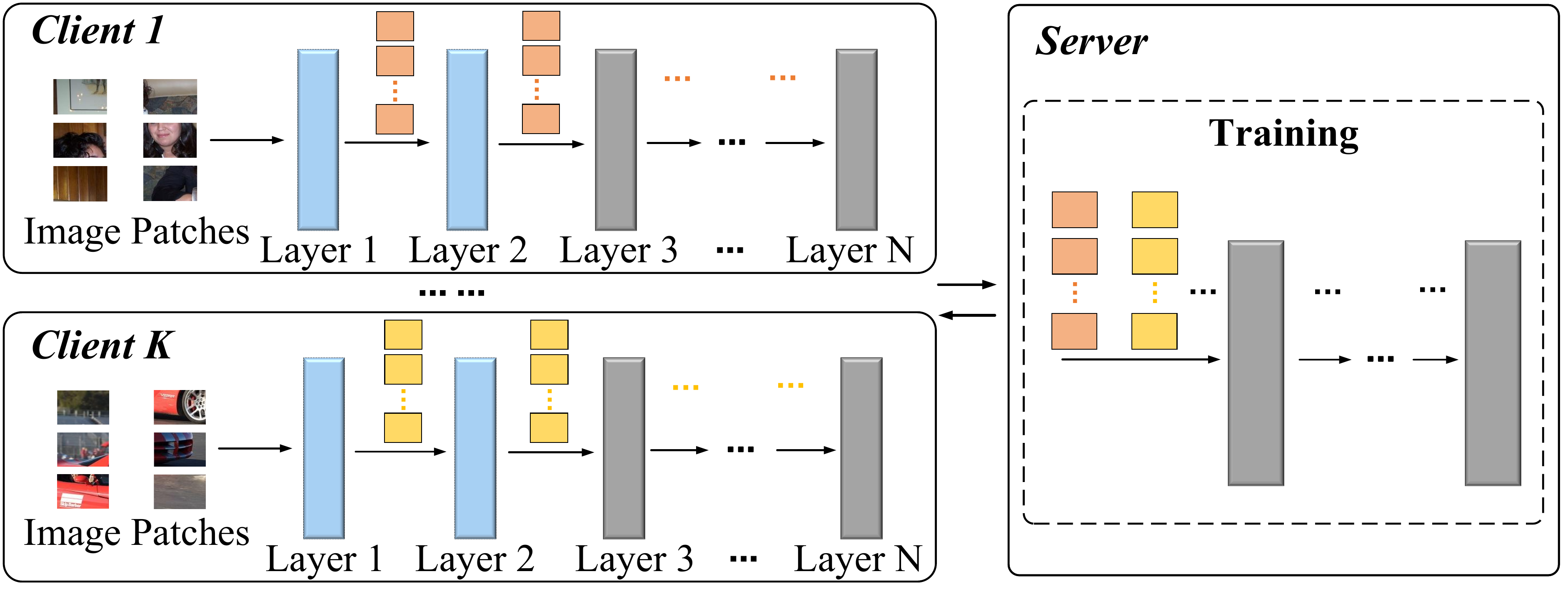}}
    \vspace{-1mm}
\caption{Comparison of our ViT-based method with previous CNN approaches in resource-limited federated learning. Previous CNN approaches include two schemes: \textit{1) Partial Parameter Aggregation} and \textit{2) Partial Parameter Training}, which rely on stage-wise model division and full-image training, restricting flexibility. In contrast, our method utilizes layer-wise division in ViTs and partial image patches for training, \textit{fundamentally reducing computational costs, time, and data privacy leakage.} }
\vspace{-5mm}
\end{figure}





In resource-constrained FL, most current research centers on CNN modals, typically split into two schemes, as illustrated in \cref{fig:p2}. 
The first scheme~\cite{diao2020heterofl,alam2022fedrolex,jia2024adaptivefl,liu2023finch}, Partial Parameter Aggregation, partitions a large server-side model into sub-models adapted to client resources for local training, then aggregates the updated parameters through a designed strategy on the server.
However, this scheme offers limited improvement in inference performance due to constraints on the size of client models.
The second scheme~\cite{he2020group,wu2024fedekt}, Partial Parameter Training, leverages intermediate features and knowledge distillation to train the majority of the parameters on the server, thereby reducing the computational load on the client. 
Nevertheless, ViTs, characterized by a larger parameter scale and patch-based processing, differ significantly from CNNs. Existing research has yet to fully optimize ViTs for resource-constrained FL.

\begin{figure}[!t]
  \centering
  \includegraphics[width=0.96\linewidth]{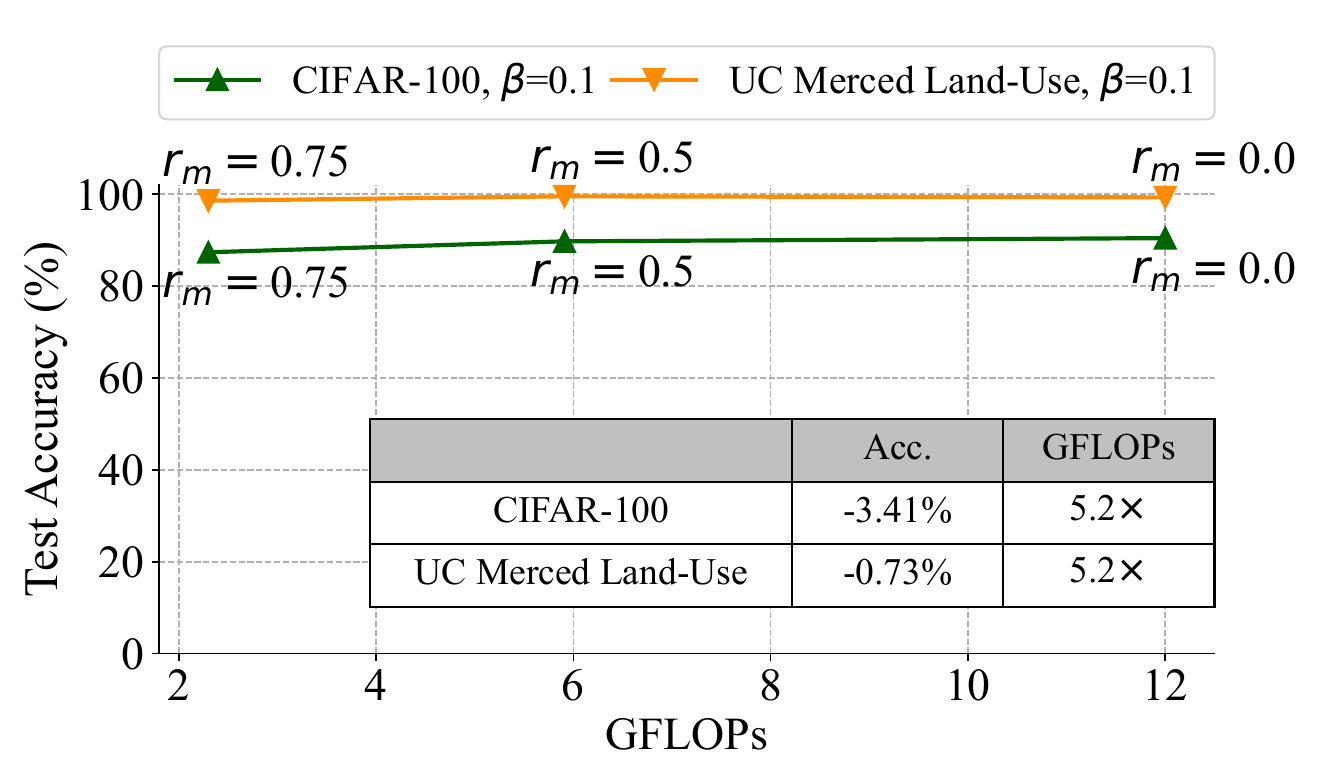}
   \vspace{-2mm}
    \caption{Impact of varying masking ratio $r_{m}$ in federated learning experiments conducted on two benchmarks \textit{without resource constraints}. Experiments are conducted under a high data heterogeneity setting with $\beta=0.1$, the concentration parameter of the Dirichlet distribution $Dir_{N}(\beta)$. Results indicate that increasing the masking ratio to 0.75 minimally affects model performance while substantially reducing training costs.}
  \label{fig:data}
  \vspace{-5mm}
\end{figure}

In this work, we explore \textit{whether the training computational costs of ViTs can be fundamentally reduced without significantly compromising FL performance}.
Recent self-supervised work has demonstrated that masked image modeling effectively learns generalizable visual representations by reconstructing randomly masked image patches~\cite{he2022masked, xie2023revealing}, highlighting the substantial redundancy in images that may be unnecessary for recognition. 
To test this hypothesis, we conduct FL experiments with \textit{no resource constraints}, using masked images to examine their impact on model performance and computational training costs.
In the experiments, images are uniformly partitioned into non-overlapping patches, with a specified ratio $r_{m}$ of patches randomly masked. Only the unmasked patches are utilized for model training. 
As illustrated in \cref{fig:data}, we conduct experiments under a challenging data heterogeneity setting with $\beta=0.1$, where $\beta$ is a concentration parameter from the Dirichlet distribution $Dir_{N}(\beta)$ in FL. 
Results indicate that varying the masking ratio has minimal impact on model accuracy but significantly reduces training computation costs. For instance, increasing $r_{m}$ from 0.00 to 0.75 reduces the computational load by up to 5.2$\times$, with only a marginal decrease in accuracy. 
These findings suggest that using masked images in FL is a promising approach for enabling efficient ViT training on resource-constrained clients.

Inspired by these observations, we propose \name{}, a hierarchical federated learning framework (as illustrated in \cref{fig:p1}) that employs masked images to efficiently train ViT models across multiple heterogeneous data on resource-constrained clients, while also enhancing privacy protection by concealing client data content.
\name{} comprises lightweight local modules on edge clients and a larger global module on the central server, designed to accommodate limited client resources. The local modules are trained on masked images. Rather than aggregating parameters from clients, the global module receives intermediate patch features from the local modules, enabling it to learn universal representations suitable for heterogeneous data.
To protect user preferences, we propose a median sampling strategy that adjusts the patch feature count for each class to the median across all classes prior to uploading, maintaining both privacy protection and training efficiency.
Our main contributions to this work are summarized as follows:
\begin{itemize}[noitemsep,topsep=0pt,leftmargin=15pt]
\item To the best of our knowledge, we present \name{}, the first federated learning framework to leverage masked images for efficiently training ViT models across multiple resource-constrained clients, while also enhancing client data content protection.

\item \name{} enables hierarchical training of all model parameters between clients and the central server, demonstrating substantial benefits for heterogeneous data. In addition, we introduce a median sampling strategy to obscure the distribution information of intermediate features before they are uploaded to the server.

\item Experiments on popular benchmarks demonstrate that \name{} reduces the computational cost of local training by as much as $5.6\times$, lowers the local training time by up to $3.1\times$, and improves the accuracy by up to 2.46\%, setting new state-of-the-art results.
\end{itemize}

%% file: 2_re.tex
\section{Related Works}

\begin{figure*}[!t]
  \centering
  \includegraphics[width=0.93\linewidth]{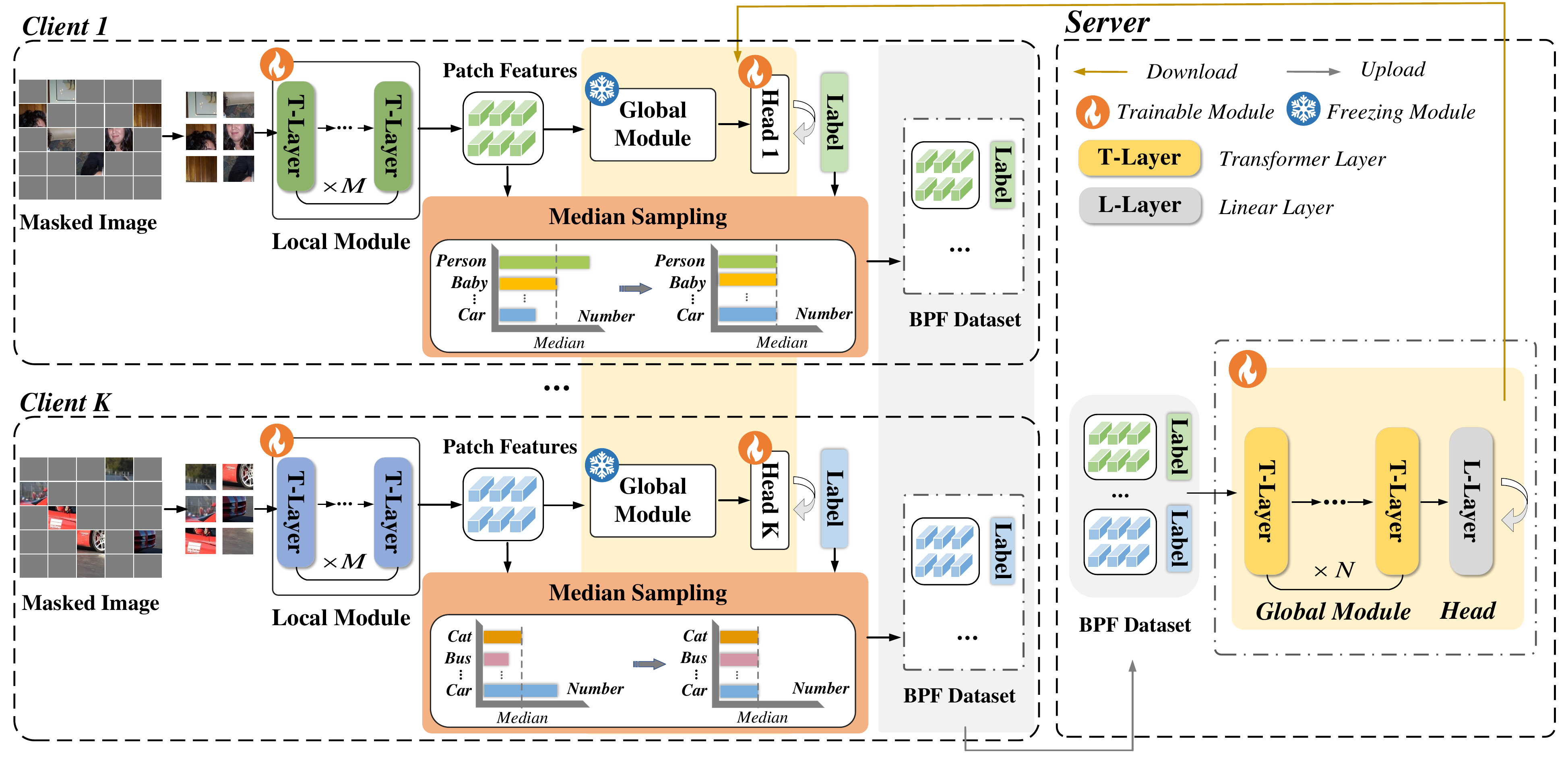}
  \vspace{-1mm}
  \caption{Overview of our hierarchical framework, \name{}, for efficient federated training of ViTs on resource-constrained clients. Local modules are trained on clients, and a shared global module is trained on a central server. Each client utilizes masked images to train its respective local module and classification head.
  Prior to uploading to the server, the proposed median sampling strategy is applied to balance patch features to match the median of class distribution, thereby enhancing client data privacy protection.
  The server trains the global module using patch features, then transmits the updated parameters back to the clients for the next training round.
  }
  \label{fig:overview}
  \vspace{-3mm}
\end{figure*}


\noindent{\textbf{Federated Learning on Resource-Constrained Clients.}}
Federated learning under client resource constraints primarily follows two paradigms: Partial Parameter Aggregation~\cite{diao2020heterofl,alam2022fedrolex,jia2024adaptivefl,liu2023finch} and Partial Parameter Training~\cite{he2020group,wu2024fedekt}.
The former partitions the global model into smaller sub-models using techniques such as design rules or pruning, enabling local training on resource-constrained clients. 
For example, HeteroFL~\cite{diao2020heterofl} randomly selects sub-models from the global model for distribution to clients, while AdaptiveFL introduces a pruning mechanism to generate heterogeneous sub-models for each client.
The latter paradigm offloads most parameter training to the server, retaining only lightweight computation on clients.  
For instance, FedGKT~\cite{he2020group} decouples the model into client-executed shallow layers and server-side deep layers, then coordinates their training via knowledge distillation.
However, existing CNN-based methods are not directly applicable to ViTs due to their larger scale and patch-based architecture.

\noindent{\textbf{Parameter-Efficient Fine-Tuning (PEFT).}} 
Deep learning has been widely applied in various scenarios such as intelligent transportation~\cite{li2021self,li2021exploiting,Yan_2025_CVPR}, cross-modal learning~\cite{Liu_2025_CVPR,zhao2025favchat,li2025uni}, privacy protection~\cite{li2023dr,li2023stprivacy}, and artificial intelligence generated content~\cite{li2024instant3d,liu2024realera,zhuang2025vistorybench}.
As these applications often rely on large models, PEFT~\cite{zaken2021bitfit,jia2022visual,hu2021lora} has emerged as an effective approach by updating only a small subset of parameters, significantly reducing storage and computational costs\cite{han2024parameter}.
Several studies~\cite{sun2022conquering,zhang2023fedpetuning} have explored using different PEFT techniques to assess performance improvements and resource savings in federated systems. 
However, the limited number of tunable parameters in PEFT inevitably constrains model adaptability, leading to suboptimal performance in federated systems.

%% file: method.tex
\section{Efficient FL with Masked Images}

\subsection{Problem Definition}
We employ supervised classification tasks distributed across $K$ clients to formulate our problem. Each client $k$ possesses a dataset $D_{k}:=(X_{k}, Y_{k})$, where $X_{k}\in \mathbb{R}^{N_{k}\times d_{k}}$ denotes the data samples and $Y_{k}\in \mathbb{R}^{N_{k}\times c_{k}}$ represents their corresponding labels. Here, $N_{k}$ represents the number of data points, $d_{k}$ denotes the input dimension, and $c_{k}$ indicates the number of classes.

\subsection{Overview}

As illustrated in \cref{fig:overview}, \name{} employs hierarchical training across clients and the central server to enable privacy-preserving and efficient collaborative learning. Each client includes a local module with $M$ Transformer layers, a shared global module with $N$ Transformer layers, and a classification head.
The local module and classification head are trained on each client with unmasked image patches $X_{p}$, enabling efficient local training and generating patch features that represent local knowledge.
To safeguard data distribution privacy, a median sampling strategy is applied on each client to create a balanced patch features (BPF) dataset before uploading to the server. 
The global module is then trained on the server using the BPF dataset from clients to effectively learn global representations for all tasks. Finally, the server transmits the updated global module parameters back to clients for next training round.

\subsection{Training with Masked Images}
To enable efficient local training on resource-constrained clients, we present a patch-wise optimization strategy. 
Firstly, each input image is divided into a sequence of regular, non-overlapping patches, which are randomly masked at a ratio $r_{m}$. The remaining unmasked patches, denoted as $X_{p}$, are then used to train our framework. We define the patch features obtained by the local module on the client $k$ $\mathcal{M}_{k}(\phi_{k}; \cdotp)$ as $H_{p}^{k}=\mathcal{M}_{k}(\phi_{k}; X_{p}^{k})$, where $X_{p}^{k}=Mask\_D(X_{k})$ and $Mask\_D(X_{k})$ is the operation of randomly masking image patches from $X_{k}$ and discarding the selected patches.
To preserve patch ordering for ViTs, the positional embeddings \cite{steiner2021train} of the remaining patches are retained.
This is inspired by the internal redundancy of images and reduces the amount of data that the model needs to process, thereby lowering computational complexity.
Additionally, these patch features $H_{p}^{k}$ make it pretty challenging to reconstruct the original images since they are encoded from a very small portion of each image, inherently providing \name{} with a content privacy advantage.
Notably, the entire images are adopted for the inference on each client.  

\subsection{Data Distribution Protection with Median Sampling} \label{sec:median_sampling}
To enhance privacy in \name{}, we propose a median sampling strategy to generate a balanced patch features dataset $D_{H}^{k}$ on each client. 
It aims to ensure that the generated patch features on each client contain an equal number of samples for each class, thereby preventing the leakage of statistical information or user preferences when uploaded to the central server. 
Imbalanced data distribution on clients is a common issue in federated learning, and the median, being less sensitive to extreme values, is well-suited for addressing this challenge. Our median sampling strategy uses the median of class sample counts on each client to differentiate between minority and majority classes. It then applies oversampling to increase samples of minority classes and downsampling to reduce samples of majority classes. Specifically, for minority class samples, all patch features generated across multiple local training epochs are retained, whereas for majority class samples, only patch features from the final epoch are preserved. 
Next, downsampling is applied to reduce the number of samples in each class to the median.
\textcolor{black}{Empirically, we find that increasing the sampling threshold adds to computation costs but does not significantly improve final performance.}

\subsection{Hierarchical Training Paradigm}

To effectively reduce the computational burden on clients without compromising performance, we propose a new hierarchical training strategy for ViTs that minimizes the number of trainable parameters on the clients. 
Further algorithmic details for the collaborative optimization procedures can be found in the supplementary materials.

\noindent{\textbf{Training on Clients.}}
On the client $k$, the local module $\mathcal{M}_{k}(\phi_{k}; \cdot)$ is responsible for mapping image patches $X_{p}$ into patch features $H_{p}$, while the global module $\mathcal{M}_{k}(w; \cdot)$ encodes $H_{p}$ into representation vectors $H_{r}$. The final classification head $\mathcal{M}_{k}(\theta_{k}; \cdot)$ transforms the representation vectors $H_{r}$ to match the number of classes.
Only the parameters of the local module and classification head are trainable, while the parameters of the global module remain frozen and are iteratively updated via downloads from the server.
For the client $k$, the loss function used in local training is defined as
{\small
\begin{equation}
  \mathcal{L}(\phi_{k},\theta_{k}) = \textstyle\sum_{i}^{c_{k}}p(y=i)log(\mathcal{M}_{k}(\phi_{k},w,\theta_{k};X_{ip}^{k})),
  \label{con:l1}
\end{equation}
}
where $c_{k}$ is the number of classes in client $k$, and $p(y=i)$ is the probability distribution of label $i$. The parameters $\phi_{k}$, $w$, $\theta_{k}$ are from the local module, global module, and classification head, respectively.
Therefore, the optimization objective is to minimize
{\small
\begin{equation}
  \min_{\phi_{k}, \theta_{k}} \mathbb{E}_{X_{ip}^{k} \sim D_{K}}[\mathcal{L} (\mathcal{M}_{k}(\phi_{k},w,\theta_{k}; X_{ip}^{k}), Y_{k})],
  \label{con:o1}
\end{equation}
}
where $\phi_{k}$ and $\theta_{k}$ are trainable.

\noindent{\textbf{Training on Server.}}
The server aggregates heterogeneous knowledge from clients to learn universal representations across diverse tasks. The global module $\mathcal{M}_{S}(w; \cdot)$ and classification head $\mathcal{M}_{S}(\theta; \cdot)$ are trained using the balanced patch features dataset uploaded from participating clients in the latest training round.
The loss function can be formulated as
{\small
\begin{equation}
  \mathcal{L}_{s}(w,\theta) = \textstyle\sum_{i}^{C}p(y=i)log(\mathcal{M}_{S}(w,\theta;H_{p}))
  \label{con:l2}
\end{equation}
}
where $C$ is the total number of classes, and $p(y=i)$ is the probability distribution of label $i$ on the data. 
The optimization objective on the server is to minimize
{\small
\begin{equation}
  \min_{w, \theta} \mathbb{E}_{H_{p} \sim D_{H}}[\mathcal{L}_{s} (\mathcal{M}_{S}(w,\theta; H_{p}), Y)],
  \label{con:o2}
\end{equation}
}
where $H_{p}$ and $Y$ represent respective patch features and labels uploaded from clients.

\begin{table*}[!t]
\centering
\resizebox{0.98\textwidth}{!}{
\begin{tabular}{@{}l c *{8}{c} @{}}
\toprule
\multirow{2.5}{*}{Dataset} & 
\multirow{2.5}{*}{$\beta$} & 
\multicolumn{8}{c}{Methods} \\ 
\cmidrule(lr){3-10}
& & Fed-Full & Fed-Head & Fed-Bias\cite{zaken2021bitfit} & Fed-Prompt\cite{jia2022visual} & Fed-LoRA\cite{hu2021lora} & FEDBFPT\cite{xin2023fedbfpt} & FedRA\cite{su2024fedra} & \textbf{Ours} \\ 
\midrule

\multirow{2}{*}{UC Merced Land-Use} 
& 0.1 & 99.31 & 76.67 & 90.48 & 86.43 & 91.19 & 90.95 & 96.43 & \textbf{98.80} \\ 
& 1.0 & 99.33 & 93.57 & 94.52 & 94.52 & 97.14 & 95.71 & 98.01  & \textbf{98.10} \\ 

\addlinespace[4pt]

\multirow{2}{*}{CIFAR-100} 
& 0.1 & 90.40 & 74.41 & 88.76 & 88.34 & 70.23 & 88.91 & 87.89 & \textbf{90.02} \\ 
& 1.0 & 92.10 & 79.57 & 90.67 & 90.45 & 84.05 & 90.37 & 90.07 & \textbf{90.81} \\ 

\addlinespace[4pt]

\multirow{2}{*}{CIFAR-10} 
& 0.1 & 98.31 & 92.27 & 97.83 & 97.99 & 96.91 & 97.99 & 98.01 & \textbf{98.12} \\ 
& 1.0 & 98.71 & 94.23 & 98.21 & 98.18 & 98.09 & 98.26 & 97.78 & \textbf{98.35} \\ 

\bottomrule
\end{tabular}}
\caption{Performance comparison between our method and state-of-the-art approaches under two data heterogeneity levels ($\beta=0.1$ and $\beta=1.0$). Notably, Fed-Full trains all parameters on clients with \textit{no resource constraints}, representing the theoretical upper bound for other methods, and is therefore excluded from the comparison. Our method, \name{}, achieves the highest accuracy across all test scenarios, demonstrating strong capability in handling highly heterogeneous data at $\beta=0.1$. Best results are highlighted in \textbf{bold}.}
\label{exp1}
\end{table*}

\subsection{Privacy \& Complexity Analysis}
\textbf{Data Content Privacy.} 
Recent studies~\cite{he2020group,wu2024fedekt} show that exchanging intermediate features during federated learning training is safer than sharing gradients. This is because attackers primarily access to evolving feature maps rather than the final, fully trained maps, making data reconstruction attacks \textcolor{black}{more challenging \cite{he2020group,yin2021see,zhao2020idlg,zhu2019deep}}. Furthermore, \name{} uploads patch features corresponding to 25\% of the image area, controlled by the masking rate $r_{m}$, which makes recovering the original image highly challenging.
To validate the effectiveness of data content privacy protection,we simulate reconstruction attacks by feeding patch features from the M ViT layers at various communication rounds into a pre-trained decoder. 
The reconstructed images remain severely distorted, indicating that even if an attacker intercepts these features, recovering the original image is highly challenging. We also observe that the degradation persists across communication rounds, suggesting consistent privacy protection throughout training. Further details are provided in \cref{sec:pa}.


\noindent\textbf{Data Distribution Privacy.} 
To protect user statistical information and preferences, our patch features are balanced via the proposed median sampling strategy on clients, ensuring an equal number of samples for each class. 
Additionally, our strategy is orthogonal to other privacy protection methods, such as Differential Privacy~\cite{dwork2008differential}, which can be seamlessly integrated into \name{} to offer enhanced protection against attacks.

\noindent \textbf{Complexity. }Given a ViT model, let $(h, w)$ represent the resolution of original image, $(p, p)$ represent the resolution of each image patch, $n=h*w/p^{2}$ be the resulting number, $d$ be the latent vector size, and $N_{T}$ represent the number of Transformer layers. To simplify the calculation, we assume that size of $Q$, $K$ and $V$ is $n \times d$.
Each client model has $N_{T}$ Transformer layers, divided into $M$ layers for local module and $N$ layers for global module. 
The model trains on $(1 - r_{m})$ of the image patches, where $r_{m}$ is the masking ratio.
The time cost for forward propagation on the client is $\mathcal{O}(5\cdot N_{T} \cdot (1-r_{m})\cdot n \cdot d^{2} + 2 \cdot N_{T}\cdot (1-r_{m})^{2} \cdot n^{2} \cdot d)$. 
As the parameters of the $N$ Transformer layers in the global module are frozen, the backward propagation time cost is $\mathcal{O}(10\cdot (N_{T}-N) \cdot (1-r_{m}) \cdot n\cdot d^{2}+4 \cdot (N_{T}-N) \cdot (1-r_{m})^{2} \cdot n^{2}\cdot d)$. Therefore, the overall time complexity in the client training stage is $\mathcal{O}( (15N_{T}- 10 N) \cdot (1-r_{m})\cdot n\cdot d^{2}+ (6N_{T}-4N) \cdot(1-r_{m})^{2}\cdot n^{2} \cdot d)$.
As $N$ approaches $N_{T}$ and $r_{m}$ approaches 1, the computational complexity of the model on the client gradually declines. Our default configurations are $N_{T}=12$, $N=10$, and $r_{m}=0.75$, substantially reducing the computational load on the client.

%% file: evaluation.tex
\section{Experiments}

\subsection{Datasets}
\label{e}

To comprehensively evaluate \name{}, we conduct experiments on two widely used federated learning datasets, CIFAR-10~\cite{krizhevsky2009learning} and CIFAR-100~\cite{krizhevsky2009learning}, as well as a more challenging datasets, UC Merced Land-Use~\cite{yang2010bag},  which is designed for scenarios with limited labeled data. 
CIFAR-10 and CIFAR-100 datasets each contain 60,000 color images. CIFAR-10 is organized into 10 classes, with 6,000 images per class (5,000 for training and 1,000 for testing), while CIFAR-100 has 100 classes, with 600 images per class (500 for training and 100 for testing).
UC Merced Land-Use dataset contains 21 land-use classes, \eg agricultural, forest, freeway, beach, and other classes, each with 100 images (80 for training and 20 for testing). 
We partition samples to all clients following a Dirichlet distribution $Dir_{N}(\beta)$ with a concentration parameter $\beta$, setting $\beta=\{0.1, 1\}$ to simulate high or low levels of heterogeneity.

\subsection{Implementations}

We use ViT-B~\cite{dosovitskiy2020image} pre-trained on ImageNet-21K~\cite{ridnik2021imagenet} as the backbone of our framework. The input images are resized to $224 \times 224$ with a patch size of $16 \times 16$. During training, data augmentation techniques such as random cropping, flipping, and brightness adjustment are applied. Following federated learning practices, we set the number of clients to 100, with a client selection ratio $P = 0.1$. The AdamW optimizer is used with an initial learning rate of $5 \times 10^{-5}$, weight decay of 0.05, and a cosine annealing learning rate schedule with warm-up. We use a batch size of 32 for both training and testing on each client. All experiments are conducted on a single NVIDIA GeForce RTX 3090 GPU. 

\begin{figure*}[!t]
  \centering
  \begin{subfigure}{0.31\linewidth}
    \includegraphics[width=\linewidth]{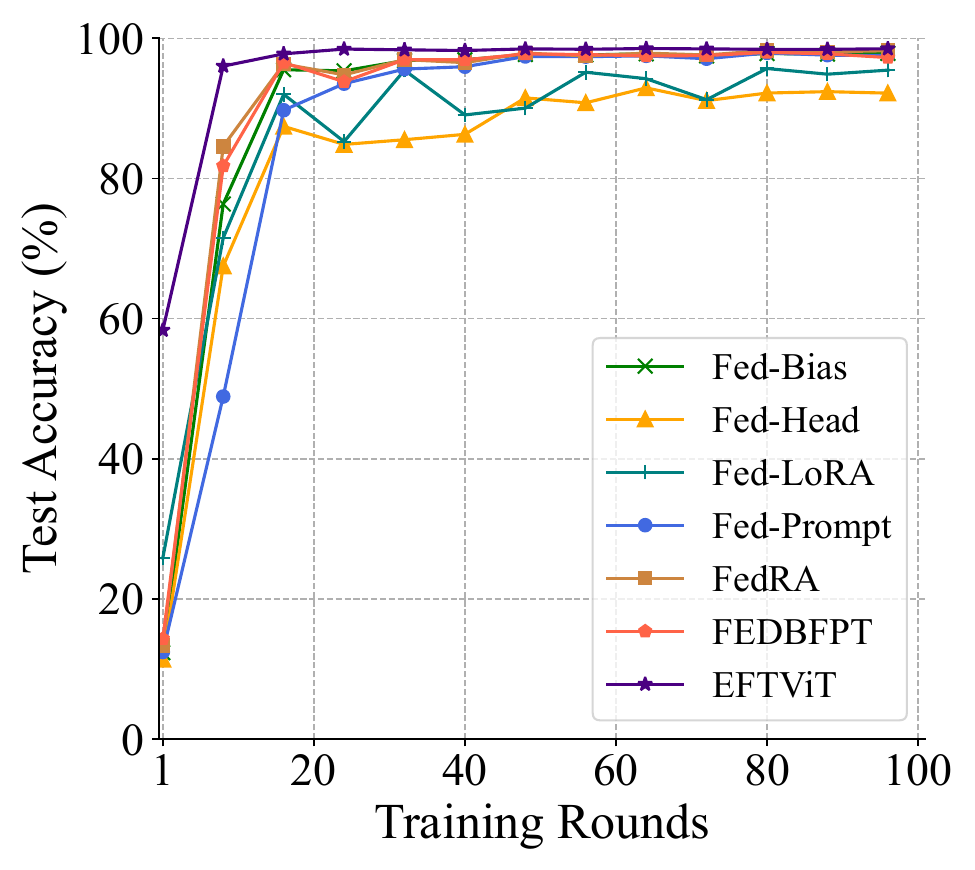}
    \caption{CIFAR-10}
  \end{subfigure}
  \hfill
  \begin{subfigure}{0.31\linewidth}
    \includegraphics[width=\linewidth]{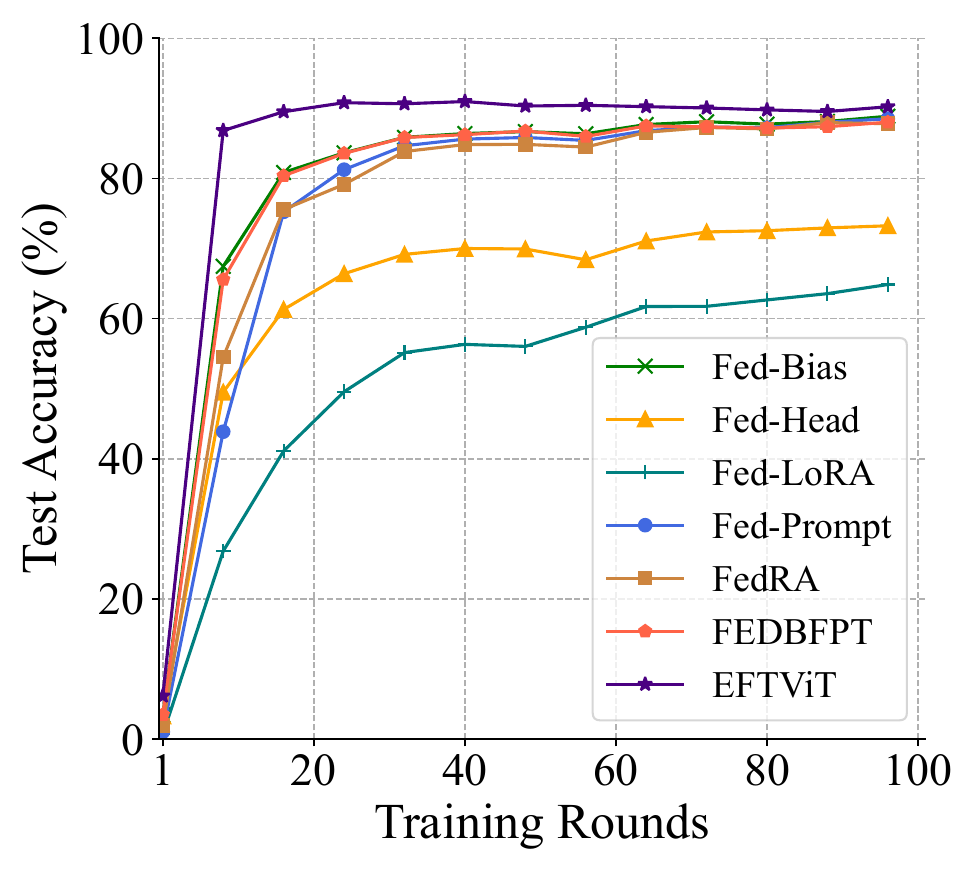}
    \caption{CIFAR-100}
  \end{subfigure}
  \hfill
  \begin{subfigure}{0.31\linewidth}
    \includegraphics[width=\linewidth]{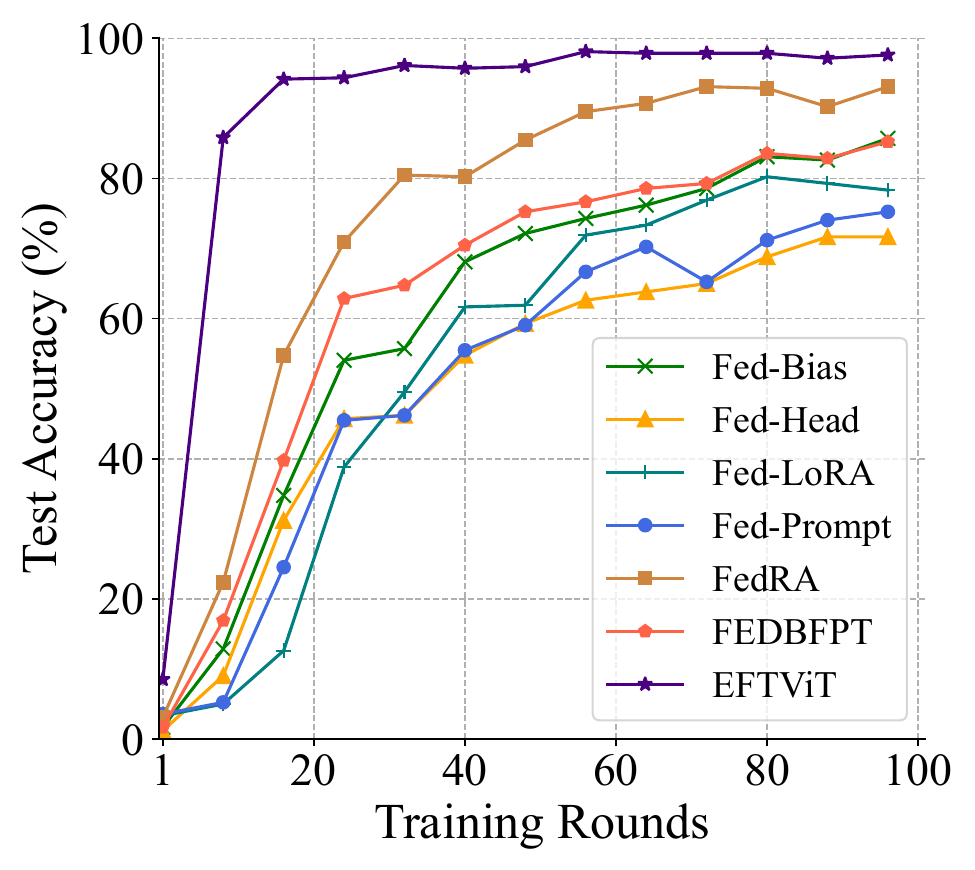}
    \caption{UC Merced Land-Use}
  \end{subfigure}
  \vspace{-1mm}
  \caption{Testing accuracy progression of \name{} and other baselines under high data heterogeneity ($\beta=0.1$) on CIFAR-10, CIFAR-100, and UC Merced Land-Use. The results show that \name{} consistently outperforms other methods throughout the training process, converging faster and more stably.}
  \label{fig:e1}
  \vspace{-2mm}
\end{figure*}

\begin{table*}[!h]
\centering
\resizebox{0.98\textwidth}{!}{
\begin{tabular}{@{}l c *{7}{c} @{}}
\toprule
\multirow{2.5}{*}{Dataset} & 
\multirow{2.5}{*}{Metric} & 
\multicolumn{7}{c}{Methods} \\ 
\cmidrule(lr){3-9}
& & Fed-Head & Fed-Bias\cite{zaken2021bitfit} & Fed-Prompt\cite{jia2022visual} & Fed-LoRA\cite{hu2021lora} & FEDBFPT\cite{xin2023fedbfpt} & FedRA\cite{su2024fedra} & \textbf{Ours} \\ 
\midrule

\multirow{2}{*}{UC Merced Land-Use} 
&  Rounds & $\infty$ & 119 & 163 & 108 & 136 &  55& 6 \\ 
& \emph{Improve $\uparrow$} & $\infty$ & \emph{19.8$\times$} & \emph{27.1$\times$} & \emph{18.0$\times$} & \emph{22.6$\times$} &  \emph{9.1$\times$} & -- \\ 

\addlinespace[4pt]

\multirow{2}{*}{CIFAR-100} 
&  Rounds & $\infty$ & 30 & 35 & $\infty$ & 31 & 49 & 4 \\ 
& \emph{Improve $\uparrow$} & $\infty$ & \emph{7.5$\times$} & \emph{8.7$\times$} & $\infty$ & \emph{7.7$\times$} & \emph{12.2$\times$} & -- \\ 

\addlinespace[4pt]

\multirow{2}{*}{CIFAR-10} 
&  Rounds & 13 & 10 & 19 & 13 & 10 & 9 & 3 \\ 
& \emph{Improve $\uparrow$} & \emph{4.3$\times$} & \emph{3.3$\times$} & \emph{6.3$\times$} & \emph{4.3$\times$} & \emph{3.3$\times$} & \emph{3.0$\times$} & -- \\ 

\bottomrule
\end{tabular}}
\caption{The number of training rounds (Rounds) required by \name{} and other baselines to reach the target accuracy of 85\%. \emph{Improve} denotes the improvement of \name{} over other methods. $\infty$ indicates the corresponding method can not reach the target accuracy. The results demonstrate that \name{} significantly reduces the convergence course.}
\label{exp3}
\end{table*}

\subsection{Comparison with State-of-the-Art Methods}

To verify the effectiveness of \name{}, we compare it with state-of-the-art resource-constrained FL methods designed for transformer-based models, as CNN-oriented approaches (e.g., quantization or pruning) are not readily adaptable to ViTs’ attention-based architecture. 
We evaluate FedRA \cite{su2024fedra} and FEDBFPT \cite{xin2023fedbfpt}. FedRA dynamically selects partial model layers for resource-constrained clients using a random allocation matrix to perform LoRA fine-tuning, thereby reducing computational overhead. FEDBFPT progressively optimizes shallower layers while selectively sampling deeper ones to lower resource consumption.
To establish additional baselines, we adapt several well-known PEFT methods to our federated learning setup: (a) \textbf{Fed-Head}: trains only the head layer parameters; (b) \textbf{Fed-Bias}: applies bias-tuning~\cite{zaken2021bitfit}, focusing on training only the bias terms; (c) \textbf{Fed-Prompt}: incorporates prompt-tuning~\cite{jia2022visual}, adding trainable prompt embeddings to the input; and (d) \textbf{Fed-LoRA}: integrates LoRA-tuning~\cite{hu2021lora} by adding the LoRA module to the query and value layers. These methods use FedAVG~\cite{mcmahan2017communication} for parameter aggregation. 
Otherwise, our method and the baseline methods share the same settings in the federated learning scenario.

\bfsection{Testing Accuracy} 
The testing results of all methods across various datasets and data heterogeneity levels are presented in \cref{exp1}. Note that Fed-Full means training all ViT parameters in clients without resource constraints, serving as a reference for the comparison. Compared with the baselines, \name{} demonstrates apparent performance gains across all scenarios. For instance, we outperform the second-best method by over 2.46\% on UC Merced Land-Use with $\beta=0.1$. Notably, our method shows consistent results in high and low data heterogeneity settings, with even better performance under higher heterogeneity. In contrast, the baseline methods degrade heavily in performance as data heterogeneity increases. These findings underscore the importance of our hierarchical training strategy in handling data heterogeneity effectively.

\bfsection{Convergence}
We report the testing accuracy changes of \name{}, FEDBFPT, and other baselines over 100 training rounds on CIFAR-10, CIFAR-100, and UC Merced Land-Use under high heterogeneity settings, as shown in \cref{fig:e1}. Our method consistently achieves the highest testing accuracy on three datasets throughout the training phase, converging faster and more stably.
To quantitatively compare convergence speed, we set a target accuracy of 85\% and record the number of training rounds (Rounds) required to reach this threshold. As shown in \cref{exp3}, \name{} significantly accelerates the convergence process, achieving 27.1$\times$ faster convergence than Fed-Prompt on the UC Merced Land-Use dataset.

\begin{table*}[!t]
\centering
\resizebox{0.96\textwidth}{!}{
\begin{tabular}{@{}c *{9}{c} @{}}
\toprule
\multirow{2.5}{*}{\begin{tabular}[c]{@{}c@{}}Metric\end{tabular}} & 
\multicolumn{8}{c}{Methods} \\ 
\cmidrule(lr){2-9}
& Fed-Full & Fed-Head & Fed-Bias\cite{zaken2021bitfit} & Fed-Prompt\cite{jia2022visual} & Fed-LoRA\cite{hu2021lora} & FEDBFPT\cite{xin2023fedbfpt} & FedRA\cite{su2024fedra} & \textbf{Ours} \\ 
\midrule

\multicolumn{1}{c}{GFLOPs} 
& 12.005 & 12.005 & 12.005 & 12.646 & 16.982 & 12.005 & 16.982 & 2.997 \\
\multicolumn{1}{c}{\emph{Improve $\uparrow$}} 
& $4.0\times$ & $4.0\times$ & $4.0\times$ & $4.2\times$ & $5.6\times$ & $4.0\times$ & $5.6\times$ & - \\ 
\midrule

\multicolumn{1}{c}{\multirow{12}{*}{\begin{tabular}[c]{@{}c@{}}
\\ \\ \\ \\ 
CTPR (s) \\
TTPR (s) \\
\emph{TT Speedup $\uparrow$} \\
Total Time (s) $\downarrow$
\end{tabular}}} 

& \multicolumn{8}{c}{UC Merced Land-Use} \\ 
& 2.619 & 0.001 & 0.004 & 0.260& 0.019&0.219 & 0.020& 1.093\\ 
& 6.887 & 4.376 & 5.506 & 5.915 & 6.087 & 5.744 & 6.436 & 2.025 \\ 
& $3.4\times$ & $2.1\times$ & $2.7\times$ & $2.9\times$ & $3.0\times$ & $2.8\times$ & $3.1\times$ & - \\ 
& 142.590 &$\infty$ &655.690 & 1006.525& 659.448& 810.968&  355.080 & 18.708\\ 

\cmidrule{2-9} 

& \multicolumn{8}{c}{CIFAR-100} \\ 
& 2.619 & 0.001 & 0.004 & 0.260& 0.019&0.219 & 0.020& 1.321\\ 
& 97.023 & 40.116 & 72.559 & 79.951 & 88.165 & 49.077 & 88.932 & 20.085 \\ 
& $4.8\times$ & $1.9\times$ & $3.6\times$ & $3.9\times$ & $4.3\times$ & $2.4\times$ & $4.4\times$ & -- \\ 
&896.778 & $\infty$ &2176.77 & 2798.285& $\infty$ & 1526.006& 4358.648 & 85.624\\ 

\cmidrule{2-9} 

& \multicolumn{8}{c}{CIFAR-10} \\ 
& 2.619 & 0.001 & 0.004 & 0.260& 0.019&0.219 & 0.020& 1.321 \\ 
& 96.378 & 42.245 & 69.782 & 75.657 & 83.657 & 51.325 & 84.312 & 18.923 \\ 
& $5.0\times$ & $2.2\times$ & $3.6\times$ & $3.9\times$ & $4.4\times$ & $2.7\times$ & $4.4\times$ & -- \\ 
& 593.982 & 549.198 & 697.860 &1442.423 & 1087.788 &515.440 &  
 758.988 & 60.723 \\ 

\bottomrule
\end{tabular}}
\caption{Training GFLOPs, communication time per round (CTPR), training time per round (TTPR), and Total Time for \name{} and other baselines on CIFAR-10, CIFAR-100, and UC Merced Land-Use. 
\emph{Improve} and \emph{TT Speedup} represent the improvement of \name{} over other baselines in GFLOPs and TTPR, respectively. The results show that \name{} significantly enhances computational efficiency.}
\label{exp2}
\end{table*}

\bfsection{Computational Efficiency}
We evaluate the client-side computational efficiency of \name{} from three perspectives: the computational cost of forward propagation during training (measured in Giga Floating-Point Operations, GFLOPs), the maximum local training time, and the total time required to achieve the target accuracy. 
The total time includes communication time, estimated under a 5G standard (500 Mbps bandwidth) to reflect realistic network conditions.
We then report the communication time per round (CTPR), training time per round (TTPR) on the client, and total time to reach 85\% accuracy (Total Time) for \name{} and other baselines across three datasets.
As shown in \cref{exp2}, \name{} significantly improves computational efficiency, achieving at least $4\times$ the efficiency of other methods in terms of GFLOPs.
For training time, \name{} reduces local training time by $3.1\times$ compared to FedRA on the UC Merced Land-Use dataset. 
Moreover, \name{} requires considerably less training time and total time than other approaches. 
These results demonstrate that our masked image and hierarchical training strategy effectively reduces client computation.

\begin{figure}[!t]
  \centering
  \includegraphics[width=0.99\linewidth]{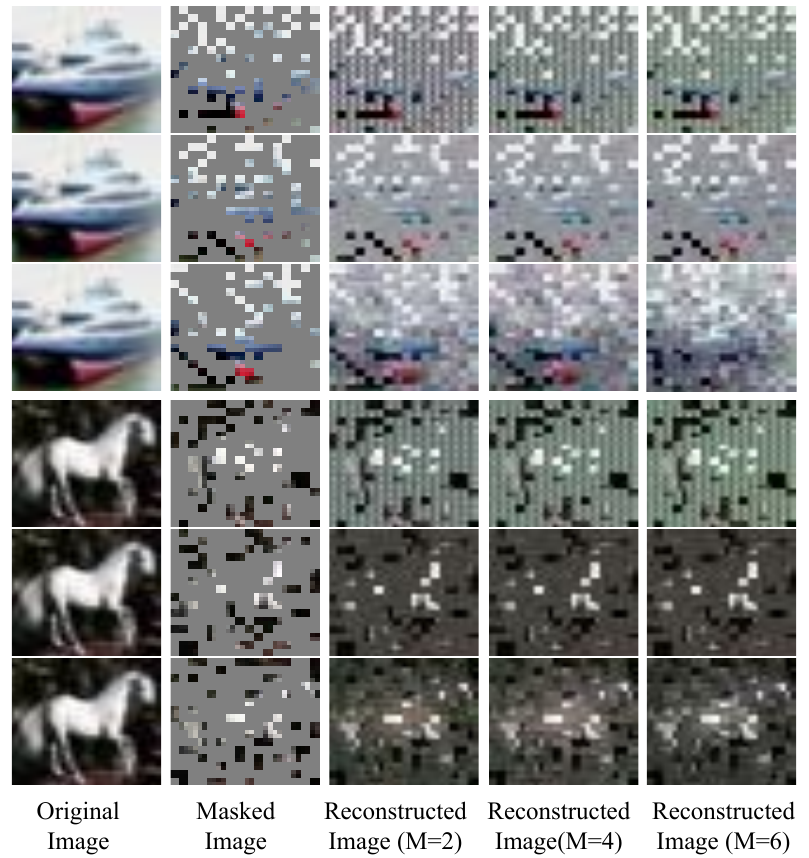}
  \caption{Visualization of reconstructed images from patch features extracted at $M$ ViT layers across different communication rounds ($T$). Each sample appears in three rows, corresponding to $T=1$, $T=10$ (near convergence in \name{}), and $T=100$. Columns show the original image, its masked version, and the image reconstructed from patch features extracted at $M\in\{2,4,6\}$. The consistently poor reconstruction quality highlights the privacy protection afforded by our method.
  }
  \label{fig:pa}
  \vspace{-6mm}
\end{figure}


\bfsection{Privacy Analysis}
\label{sec:pa}
To evaluate the privacy protection provided by transmitting patch features in our \name{}, we reconstruct masked patches by feeding patch features extracted from $M\in\{2,4,6\}$ ViT layers at various communication rounds ($T$) into a pre-trained decoder. 
\cref{fig:pa} presents representative results for two image samples, each with three rows corresponding to $T=1$, $T=10$ (near-convergence in \name{}), and $T=100$. 
The poor reconstruction quality indicates that even if an attacker intercepts these patch features, recovering the original image would be extremely difficult, highlighting the privacy advantages of \name{}.

\subsection{Ablation Study}
We conduct extensive ablation experiments to investigate the key components of our approach.

\begin{figure}[!h]
  \centering
  \includegraphics[width=0.98\linewidth]{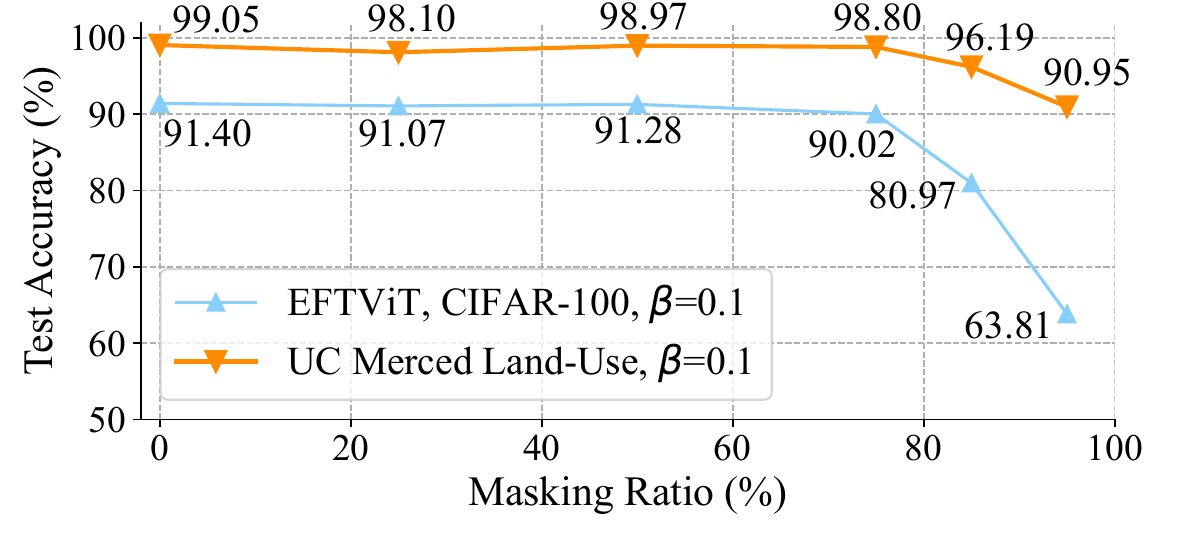}
  \vspace{-2mm}
  \caption{Accuracy changes of \name{} with varying masking ratio on CIFAR-100 at $\beta=0.1$. Our \name{} is shown to work on a wide masking ratio range. The testing accuracy fluctuates marginally and is larger than 90\% when the masking ratio increases from 0 to 75\%. However, the performance deteriorates apparently when the masking ratio exceeds 75\%.}
  \label{fig:ee3}
  \vspace{-3mm}
\end{figure}

\bfsection{Effect of Masking Ratio}
The masking ratio $r_{m}$ determines the number of masked image patches. A smaller $r_{m}$ reduces the amount of input data, thus lowering computational requirements during model training. \cref{tab:rm1} provides the GFLOPs for various masking rates, demonstrating that increasing the masking ratio significantly reduces GFLOPs. However, increasing the masking ratio also affects overall performance. We evaluate the effect of different masking rates for \name{}.
\cref{fig:ee3} shows the results of \name{} with varying masking ratios on CIFAR-100 and UC Merced Land-Use at $\beta=0.1$. Results indicate that \name{} can support a wide masking ratio range. When the masking ratio increases from 0\% to 75\%, the accuracy remains larger than 90\%. However, the performance decreases heavily when the masking ratio exceeds 75\%. Therefore, we select a masking ratio of 75\% to strike a balance between accuracy and computational efficiency.

\begin{table}[!t]
\centering
\resizebox{0.38\textwidth}{!}{
\begin{tabular}{cccccc}
\toprule
$r_{m}$     & 0.00   & 0.25  & 0.50  & 0.75  & 0.95  \\ 
\cmidrule{2-6}
GFLOPs & 12.005 & 8.914 & 5.911 & 2.997 & 0.684 \\ \bottomrule

\end{tabular}}
\caption{GFLOPs calculated by different $r_m$. GFLOPs decrease significantly as the masking rate increases.
}
\label{tab:rm1}
\vspace{-2mm}
\end{table}


\bfsection{Effect of Layer Number $M$ in Local Module}
The layer number $M$ determines the trainable parameter division between clients and the server, affecting the computational load of clients and final performance. \cref{tab:t5} presents the number of trainable parameters (Params) in each client and the corresponding accuracy achieved by the model for different values of $M$.
The results show that $M$ has minimal impact on the testing accuracy, showcasing the superior robustness of our \name{} \wrt client resources. Given the higher computational cost of a large $M$ on clients and the accuracy decrease, we select $M=2$ as the default setting.

\begin{figure}[!t]
  \centering
  \includegraphics[width=0.99\linewidth]{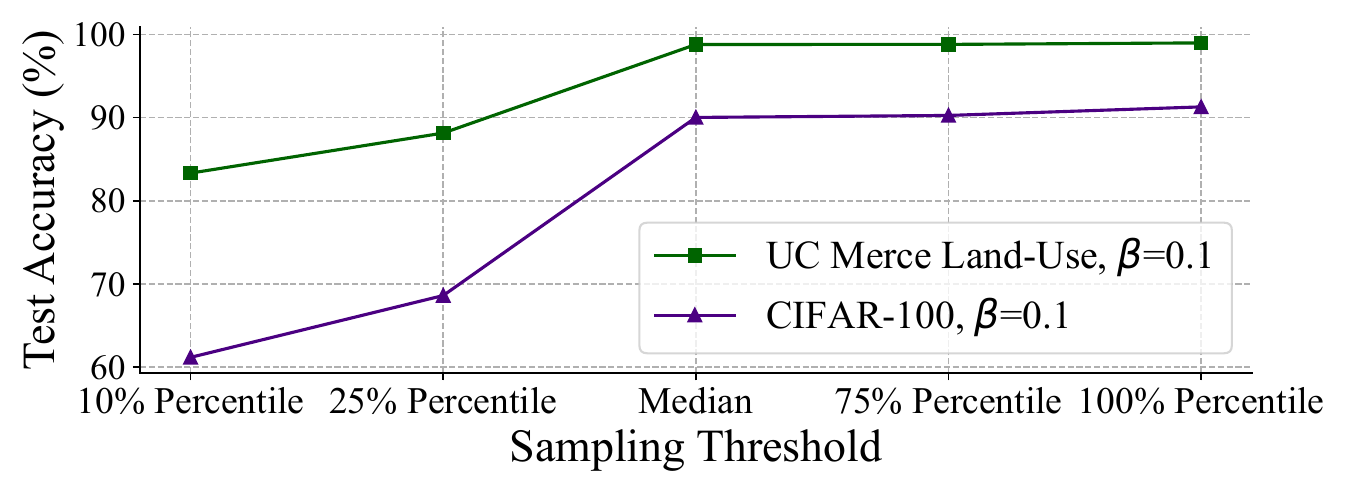}
  \vspace{-2mm}
  \caption{Accuracy of our approach with different sampling thresholds for balancing patch features in \cref{sec:median_sampling} on three datasets at $\beta=0.1$. Results suggest that increasing the threshold does not significantly enhance performance.}
  \label{fig:ee4}
  \vspace{-3mm}
\end{figure}

\bfsection{Effect of Sampling Threshold}
As elaborated in \cref{sec:median_sampling}, the sampling threshold determines the number of balanced patch features to upload for server training. Therefore, a higher threshold increases the training cost on the server.
We investigate the impact of utilizing median or higher sampling thresholds in \name{}, as shown in \cref{fig:ee4}. Results indicate that increasing the threshold provides minimal performance improvements. 
To enhance the computational efficiency on the server, we select the median as the threshold.

\begin{table}[!t]
\centering
\resizebox{0.46\textwidth}{!}{
\begin{tabular}{ccccc}
\toprule
\multicolumn{1}{c}{\multirow{2}{*}[-1.5ex]{$M$}} & \multicolumn{1}{c}{\multirow{2}{*}[-1.5ex]{Params}} & \multicolumn{3}{c}{Accuracy~(\%)}                                                           \\ \cmidrule{3-5} 
\multicolumn{1}{c}{}                   & \multicolumn{1}{c}{}                        & 
CIFAR-10 & CIFAR-100 & UC Merce Land-Use\\ \midrule
2    & 14.23M     & 98.35   & 90.81     & 98.80      \\
4    & 27.82M     & 97.56    & 90.14     & 97.85     \\ 
6    & 41.34M    & 97.98    &  89.37    &  97.38    \\ 
\bottomrule
\end{tabular}}
\caption{Accuracy and number of trainable parameters (Params) on each client for different layer numbers $M$. Results demonstrate that our \name{} has superior robustness \wrt client resources. }
\label{tab:t5}
\vspace{-2mm}
\end{table}


%% file: conclusion.tex
\section{Conclusion}
\label{cd}

In this work, we propose a hierarchical federated framework, \name{}, designed for efficient training on resource-constrained clients and handling heterogeneous data effectively. \name{} reduces client computation by leveraging masked images with an appropriate masking ratio, which minimizes performance degradation while significantly lowering computational overhead by exploiting redundancy in image information. The masked images can also prevent the data content leakage from uploaded local features. Additionally, the hierarchical training strategy, which splits parameter training between the client and server, achieves full parameter optimization and improves performance on heterogeneous data across multiple clients. Finally, \name{} incorporates a median sampling strategy to protect user data distribution, ensuring privacy while maintaining robust performance. 
The extensive experiments on three benchmarks demonstrate that \name{} significantly improves classification accuracy, reduces client training computational costs and time by large margins.


%% file: Acknowledgements.tex

\vspace{1em}
\bfsection{Acknowledgements} This research was supported by the Guangdong Basic and Applied Basic Research Foundation
(No. 2024A1515011774) and the National Natural Science Foundation of China (62302385).